\documentclass[draftcls,onecolumn]{IEEEtran}
\ifCLASSOPTIONcompsoc
  \usepackage[nocompress]{cite}
\else
  \usepackage{cite}
\fi

\ifCLASSINFOpdf
  \usepackage[pdftex]{graphicx}
\else
\fi
\usepackage[cmex10]{amsmath}
\usepackage{amsfonts} 
\usepackage{array}
\usepackage{fixltx2e}

\usepackage{stfloats}
\usepackage{url}

\begin{document}
%
\title{Highly comparative feature-based time-series classification}
%
%
%

\author{Ben~D.~Fulcher and Nick~S.~Jones%
\IEEEcompsocitemizethanks{\IEEEcompsocthanksitem B. D. Fulcher is with the Department of Physics, University of Oxford, UK and the Department of Mathematics, Imperial College London, UK.\protect\\
Email: ben.d.fulcher@gmail.com
\IEEEcompsocthanksitem N. S. Jones is with the Department of Mathematics, Imperial College London, UK.}%
\thanks{}}
\markboth{IEEE Transactions in Knowledge and Data Engineering}%
{Fulcher and Jones: Highly comparative, feature-based time-series classification}
%



\maketitle

\begin{abstract}
A highly comparative, feature-based approach to time series classification is introduced that uses an extensive database of algorithms to extract thousands of interpretable features from time series.
These features are derived from across the scientific time-series analysis literature, and include summaries of time series in terms of their correlation structure, distribution, entropy, stationarity, scaling properties, and fits to a range of time-series models.
After computing thousands of features for each time series in a training set, those that are most informative of the class structure are selected using greedy forward feature selection with a linear classifier.
The resulting feature-based classifiers automatically learn the differences between classes using a reduced number of time-series properties, and circumvent the need to calculate distances between time series.
Representing time series in this way results in orders of magnitude of dimensionality reduction, allowing the method to perform well on very large datasets containing long time series or time series of different lengths.
For many of the datasets studied, classification performance exceeded that of conventional instance-based classifiers, including one nearest neighbor classifiers using Euclidean distances and dynamic time warping and, most importantly, the features selected provide an understanding of the properties of the dataset, insight that can guide further scientific investigation.
\end{abstract}


%
\IEEEpeerreviewmaketitle

\section{Introduction}
%
%
%
%
\IEEEPARstart{T}{ime} series, measurements of some quantity taken over time, are measured and analyzed across the scientific disciplines, including human heart beats in medicine, cosmic rays in astrophysics, rates of inflation in economics, air temperatures in climate science, and sets of ordinary differential equations in mathematics.
The problem of extracting useful information from time series has similarly been treated in a variety of ways, including an analysis of the distribution, correlation structures, measures of entropy or complexity, stationarity estimates, fits to various linear and nonlinear time-series models, and quantities derived from the physical nonlinear time-series analysis literature.
However, this broad range of scientific methods for understanding the properties and dynamics of time series has received less attention in the temporal data mining literature, which treats large databases of time series, typically with the aim of either clustering or classifying the data \cite{Keogh03, Gan07, Mitsa10}.
Instead, the problem of time-series clustering and classification has conventionally been addressed by defining a distance metric between time series that involves comparing the sequential values directly.
Using an extensive database of algorithms for measuring thousands of different time-series properties (developed in previous work \cite{Fulcher13}), here we show that general feature-based representations of time series can be used to tackle classification problems in time-series data mining.
The approach is clearly important for many applications across the quantitative sciences where unprecedented amounts of data are being generated and stored, and also for applications in industry (e.g., classifying anomalies on a production line), finance (e.g., characterizing share price fluctuations), business (e.g., detecting fraudulent credit card transactions), surveillance (e.g., analyzing various sensor recordings), and medicine (e.g., diagnosing heart beat recordings).
Two main challenges of time-series classification are typically: (i) selecting an appropriate {\it representation} of the time series, and (ii) selecting a suitable measure of dissimilarity or {\it distance} between time series \cite{Wang12}.
The literature on representations and distance measures for time-series clustering and classification is extensive \cite{Wang12, Keogh03, Liao05}.
Perhaps the most straightforward representation of a time series is its time-domain form, then distances between time series relate to differences between the time-ordered measurements themselves.
When short time series encode meaningful patterns that need to be compared, new time series can be classified by matching them to similar instances of time series with a known classification.
This type of problem has traditionally been the focus of the time series data mining community \cite{Keogh03, Wang12}, and we refer to this approach as {\it instance-based} classification.
An alternative approach involves representing time series using a set of derived properties, or features, and thereby transforming the temporal problem to a static one \cite{Wang08}.
A very simple example involves representing a time series using just its mean and variance, thereby transforming time-series objects of any length into short vectors that encapsulate these two properties.
Here we introduce an automated method for producing such {\it feature-based} representations of time series using a large database of time-series features.
We note that not all methods fit neatly into these two categories of instance-based and feature-based classification.
For example, time-series shapelets \cite{Ye09, Rakthanmanon13} classify new time series according to the minimum distance of particular time-series subsequences (or `shapelets') to that time series.
Although this method uses distances calculated in the time-domain as a basis for classification (not features), new time series do not need to be compared to a large number of training instances (as in instance-based classification).
In this paper we focus on a comparison between instance-based classification and our feature-based classifiers.

Feature-based representations of time series are used across science, but are typically applied to longer time series corresponding to streams of data (such as extended medical or speech recordings) rather than the short pattern-like time series typically studied in temporal data mining.
Nevertheless, some feature-based representations of shorter time series have been explored previously: for example, Nanopoulos et al. used the mean, standard deviation, skewness, and kurtosis of the time series and its successive increments to represent and classify control chart patterns \cite{Nanopoulos01}, M\"orchen used features derived from wavelet and Fourier transforms of a range of time-series datasets to classify them \cite{Morchen03}, Wang et al. introduced a set of thirteen features that contains measures of trend, seasonality, periodicity, serial correlation, skewness, kurtosis, chaos, nonlinearity, and self-similarity to represent time series \cite{Wang06}, an approach that has since been extended to multivariate time series \cite{Wang07}, and Deng et al. used measures of mean, spread, and trend in local time-series intervals to classify different types of time series \cite{Deng13}.
As with the choice of representations and distance metrics for time series, features for time-series classification problems are usually selected manually by a researcher for a given dataset.
However, it is not obvious that the features selected by a given researcher will be the best features with which to distinguish the known data classes---perhaps simpler alternatives exist with better classification performance?
Furthermore, for many applications, the mechanisms underlying the data are not well understood, making it difficult to develop a well-motivated set of features for classification.

In this work, we automate the selection of features for time-series classification by computing thousands of features from across the scientific time-series analysis literature and then selecting those with the best performance.
The classifier is thus selected according to the structure of the data rather than the methodological preference of the researcher, with different features selected for different types of problems: e.g., we might discover that the variance of time series distinguishes classes for one type of problem, but their entropy may be important for another.
The process is completely data-driven and does not require any knowledge of the dynamical mechanisms underlying the time series or how they were measured.
We describe our method as `highly comparative' \cite{Fulcher13} and draw an analogy to the DNA microarray, which compares large numbers of gene expression profiles simultaneously to determine those genes that are most predictive of a target condition; here, we compare thousands of features to determine those that are most suited to a given time-series classification task.
As well as producing useful classifiers, the features selected in this way highlight the types of properties that are informative of the class structure in the dataset and hence can provide new understanding.

\section{Data and Methods} \label{sec:DM_methods}
Central to our approach is the ability to represent time series using a large and diverse set of their measured properties.
In this section, we describe how this representation is constructed and how it forms a basis for classification.
In Sec.~\ref{sec:DM_data}, the datasets analyzed in this work are introduced.
The feature-vector representation of time series is then discussed in Sec.~\ref{sec:DM_feature_vector}, and the methodology used to perform feature selection and classification is described in Sec.~\ref{sec:DM_feature_selection}.



\subsection{Data} \label{sec:DM_data}

The twenty datasets analyzed in this work are obtained from {\it The UCR Time Series Classification/Clustering Homepage} \cite{Keogh06a}.
All datasets are of labeled, univariate time series and all time series in each dataset have the same length.
Note that this resource has since (late in 2011) been updated to include an additional twenty-five datasets \cite{Keogh11}, which are not analyzed here.
The datasets (which are listed in Tab.~\ref{tab:cfnresults} and described in more detail in Supplementary Table~I), span a range of: 
(i) time-series lengths, $N$, from $N =$ 60 for the Synthetic Control dataset, to $N =$ 637 samples for Lightning (two);
(ii) dataset sizes, from a number of training ($n_\mathrm{train}$) and test ($n_\mathrm{test}$) time series of $n_\mathrm{train} =$ 28 and $n_\mathrm{test} =$ 28 for Coffee, to $n_\mathrm{train} =$ 1\,000 and $n_\mathrm{test} =$ 6\,164 for Wafer;
and (iii) number of classes, $n_\mathrm{classes}$, from $n_\mathrm{classes} =$ 2 for Gun point, to $n_\mathrm{classes} =$ 50 for 50 Words.
The datasets are derived from a broad range of systems: including measurements of a vacuum-chamber sensor during the etch process of silicon wafer manufacture (Wafer \cite{Olszweski01}), spectrograms of different types of lightning strikes (Lightning \cite{Eads02}), the shapes of Swedish leaves (Swedish Leaf \cite{Soderkvist01}), and yoga poses (Yoga \cite{Wei06}).
All the data is used exactly as obtained from the {\it UCR} source \cite{Keogh06a}, without any preprocessing and using the specified partitions of each dataset into training and test portions.
The sensitivity of our results to different such partitions is compared for all datasets in Supplementary Table II; test set classification rates are mostly similar to those for the given partitions. 
We present only results for the specified partitions throughout the main text to aid comparison with other studies.

\begin{figure*}[t]
	\centering
		\includegraphics[width = 16.8cm]{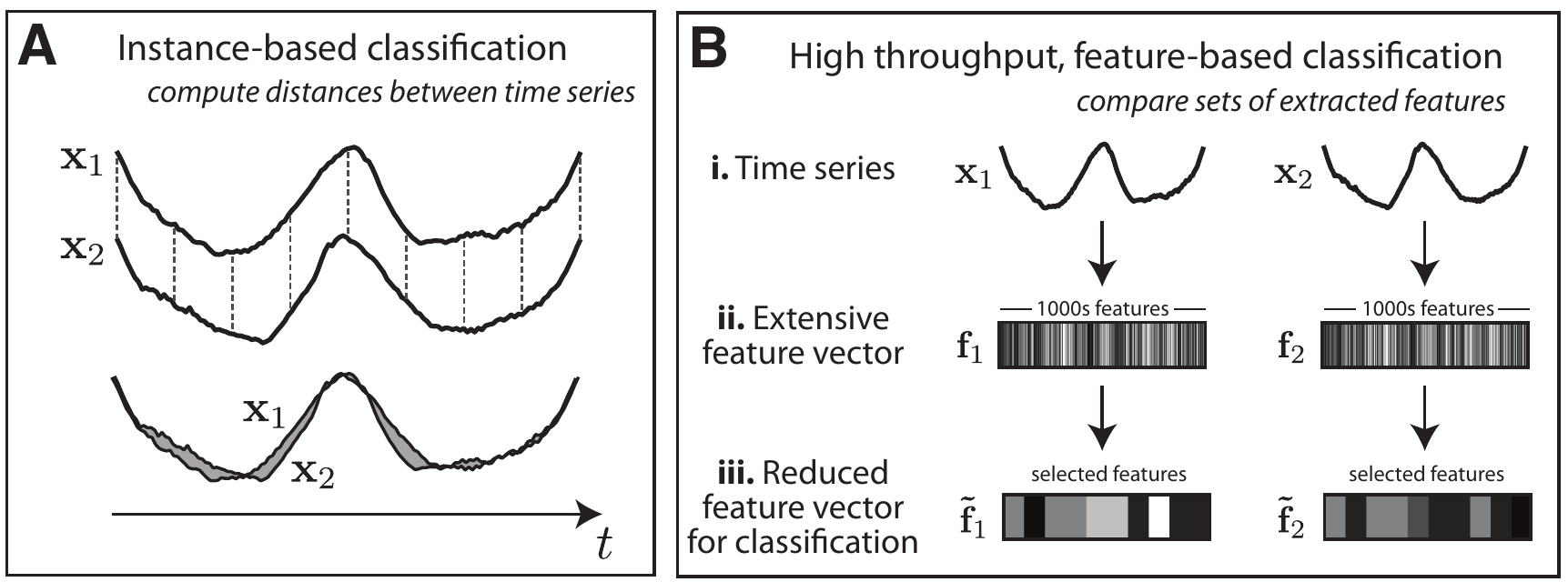}
	\caption{
	\textbf{A visual comparison of two different approaches to comparing time series that form the basis for instance-based and feature-based classification.}
	\textbf{A} Instance-based classification involves measuring the distance between pairs of time series represented as an ordered set of measurements in the time domain.
	In the upper portion of the plot, the two time series, $\mathbf{x}_1$ and $\mathbf{x}_2$, are offset vertically for clarity, but are overlapping in the lower plot, where shading has been used to illustrate the distance between $\mathbf{x}_1$ and $\mathbf{x}_2$.
	\textbf{B} An alternative approach, that forms the focus of this work, involves representing time series using a set of features that summarize their properties.
	Each time series, $\mathbf{x}$, is analyzed by computing a large number of time-series analysis algorithms, yielding an extensive set of features, $\mathbf{f}$, that encapsulates a broad range of its properties.
	Using the structure of a labeled training dataset, these features can then be filtered to produce a reduced, feature-based representation of the time series, $\mathbf{\tilde{f}}$, and a classification rule learned on $\mathbf{\tilde{f}}$ can then be used to classify new time series.
	In the figure, feature vectors are normalized and represented using a grayscale color map, where black represents low values and white represents high values of a given feature.
	}
	\label{fig:method_illustration}
\end{figure*}

\subsection{Feature vector representation} \label{sec:DM_feature_vector}

Feature-based representations of time series are constructed using an extensive database of over 9\,000 time-series analysis operations developed in previous work \cite{Fulcher13}.
The operations quantify a wide range of time-series properties, including basic statistics of the distribution of time-series values (e.g., location, spread, Gaussianity, outlier properties), linear correlations (e.g., autocorrelations, features of the power spectrum), stationarity (e.g., StatAv, sliding window measures, prediction errors), information theoretic and entropy/complexity measures (e.g., auto-mutual information, Approximate Entropy, Lempel-Ziv complexity), methods from the physical nonlinear time-series analysis literature (e.g., correlation dimension, Lyapunov exponent estimates, surrogate data analysis), linear and nonlinear model fits [e.g., goodness of fit estimates and parameter values from autoregressive moving average (ARMA), Gaussian Process, and generalized autoregressive conditional heteroskedasticity (GARCH) models], and others (e.g., wavelet methods, properties of networks derived from time series, etc.) \cite{Fulcher13}.
All of these different types of analysis methods are encoded algorithmically as operations.
Each operation, $\rho$, is an algorithm that takes a time series, $\mathbf{x} = (x_1,x_2,...,x_N)$, as input, and outputs a single real number, i.e., $\rho: \mathbb{R}^N \rightarrow \mathbb{R}$.
We refer to the output of an operation as a `feature' throughout this work.
All calculations are performed using Matlab 2011a (a product of The MathWorks, Natick, MA).
Although we use over 9\,000 operations, many groups of operations result from using different input parameters to the same type of time-series method (e.g., autocorrelations at different time lags), making the number of conceptually-distinct operations significantly smaller: approximately 1\,000 according to one estimate \cite{Fulcher13}.
The Matlab code for all the operations used in this work can be explored and downloaded at \url{www.comp-engine.org/timeseries}.



Differences between instance-based time-series classification, where distances are calculated between the ordered values of the time series, and feature-based time-series classification, which learns a classifier using a set of features extracted from the time series, are illustrated in Fig.~\ref{fig:method_illustration}.
Although the simplest `lock step' distance measure \cite{Wang12} is depicted in Fig.~\ref{fig:method_illustration}A, more complex choices, such as dynamic time warping (DTW) \cite{Berndt94}, can accommodate unaligned patterns in the time series, for example \cite{Wang12}.
The method proposed here is depicted in Fig.~\ref{fig:method_illustration}B, and involves representing time series as extensive feature vectors, $\mathbf{f}$, which can be used as a basis for selecting a reduced number of informative features, $\mathbf{\tilde{f}}$, for classification.
Although we focus on classification in this work, we note that dimensionality reduction techniques, such as principal components analysis, can be applied to the full feature vector, $\mathbf{f}$, which can yield meaningful lower-dimensional representations of time-series datasets that can be used for clustering, as demonstrated in previous work \cite{Fulcher13}, and illustrated briefly for the Swedish Leaf dataset in Supplementary Fig.~1. 

In some rare cases, an operation may output a `special value', such as an infinity or imaginary number, or it may not be appropriate to apply it to a given time series, e.g., when a time series is too short, or when a positive-only distribution is being fit to data that is not positive.
Indeed, many of the operations used here were designed to measure complex structure in long time-series recordings, such as the physical nonlinear time-series analysis literature and some information theoretic measures, that require many thousands of points to produce a robust estimate of that feature, rather than the short time-series patterns of 100s of points or less analyzed here.
In this work, we filtered out all operations that produced any special values on a dataset prior to performing any analysis.
After removing these operations, between 6\,220 and 7\,684 valid operations remained for the datasets studied here.

\subsection{Feature selection and classification} \label{sec:DM_feature_selection}
Feature selection is used to select a reduced set of features, $\tilde{\mathbf{f}} = \{\tilde{f}_i\}$, from a large initial set of thousands, $\mathbf{f} = \{f_i\}$, with the aim of producing such a set, $\tilde{\mathbf{f}}$, that best contributes to distinguishing a known classification of the time series.
Many methods have been developed for performing feature selection \cite{Jain00, Guyon03, Guyon07}, including the {\it Lasso} \cite{Tibshirani96} and recursive feature elimination \cite{Guyon02}.
In this work we use a simple and interpretable method: greedy forward feature selection, which grows a set of important features incrementally by optimizing the linear classification rate on the training data \cite{Hastie09}.
Although better performance could be achieved using more complex feature selection and classification methods, we value transparency over sophistication to demonstrate our approach here.
The greedy forward feature selection algorithm is as follows: (i) Using a given classifier, the classification rates of all individual features, $f_i$, are calculated and the feature with the highest classification rate is selected as the first feature in the reduced set, $\tilde{f}_1$. (ii) The classification rates of all features in combination with $\tilde{f}_1$ are calculated and the feature that, in combination with $\tilde{f}_1$, produces a classifier with the highest classification rate is chosen next as $\tilde{f}_2$. (iii) The procedure is repeated, choosing the operation that provides the greatest improvement in classification rate at each iteration until a termination criterion is reached, yielding a reduced set of $m$ features: $\mathbf{\tilde{f}} = \{\tilde{f}_1, \tilde{f}_2, ..., \tilde{f}_m\}$.
For iterations at which multiple features produce equally good classification rates, one of them is selected at random.
Feature selection is terminated at the point at which the improvement in the training set classification rate upon adding an additional feature drops below 3\%, or when the training set misclassification rate drops to 0\% (after which no further improvement is possible).
Our results are not highly sensitive to setting this threshold at 3\%; this sensitivity is examined in Supplementary Fig.~3. 

To determine the classification rate of each feature (or combination of features), we use a linear discriminant classifier, implemented using the \textbf{classify} function from Matlab's Statistics Toolbox, which fits a multivariate normal density to each class using a pooled estimate of covariance.
Because the linear discriminant is so simple, over-fitting to the training set is not problematic, and we found that using 10-fold cross validation within the training set produced similar overall results.
Cross validation can also be difficult to apply to some datasets studied here, which can have as few as a single training example for a given class.
For datasets with more than two classes, linear classification boundaries are constructed between all pairs of classes and new time series are classified by evaluating all classification rules and then assigning the new time series to the class with the most `votes' from this procedure.
The performance of our linear feature-based classifier is compared to three different instance-based classifiers, which are labeled as:
(i) `Euclidean 1-NN', a 1-NN classifier using the Euclidean distance,
(ii) `DTW 1-NN', a 1-NN classifier using a dynamic time warping distance, and
(iii) `DTW 1-NN (best warping window, $r$)', a 1-NN classifier using a dynamic time warping distance with a warping window learned using the Sakoe-Chiba Band (cf. \cite{Ratanamahatana04}).
These results were obtained from {\it The UCR Time Series Classification/Clustering Homepage} \cite{Keogh06a}.
Results using a 1-NN classifier with Euclidean distances were verified by us and were consistent with the UCR source \cite{Keogh06a}.


\section{Results} \label{sec:results}
In this section, we demonstrate our highly comparative, feature-based approach to time-series classification.
In Sec.~\ref{sec:particular_datasets} we illustrate the method using selected datasets, in Sec.~\ref{sec:all_results} we compare the results to instance-based classification methods across all twenty datasets, and in Sec.~\ref{sec:computational_issues} we discuss the computational complexity of our method.

\begin{figure*}[t]
	\centering
		\includegraphics[width = 17.5cm]{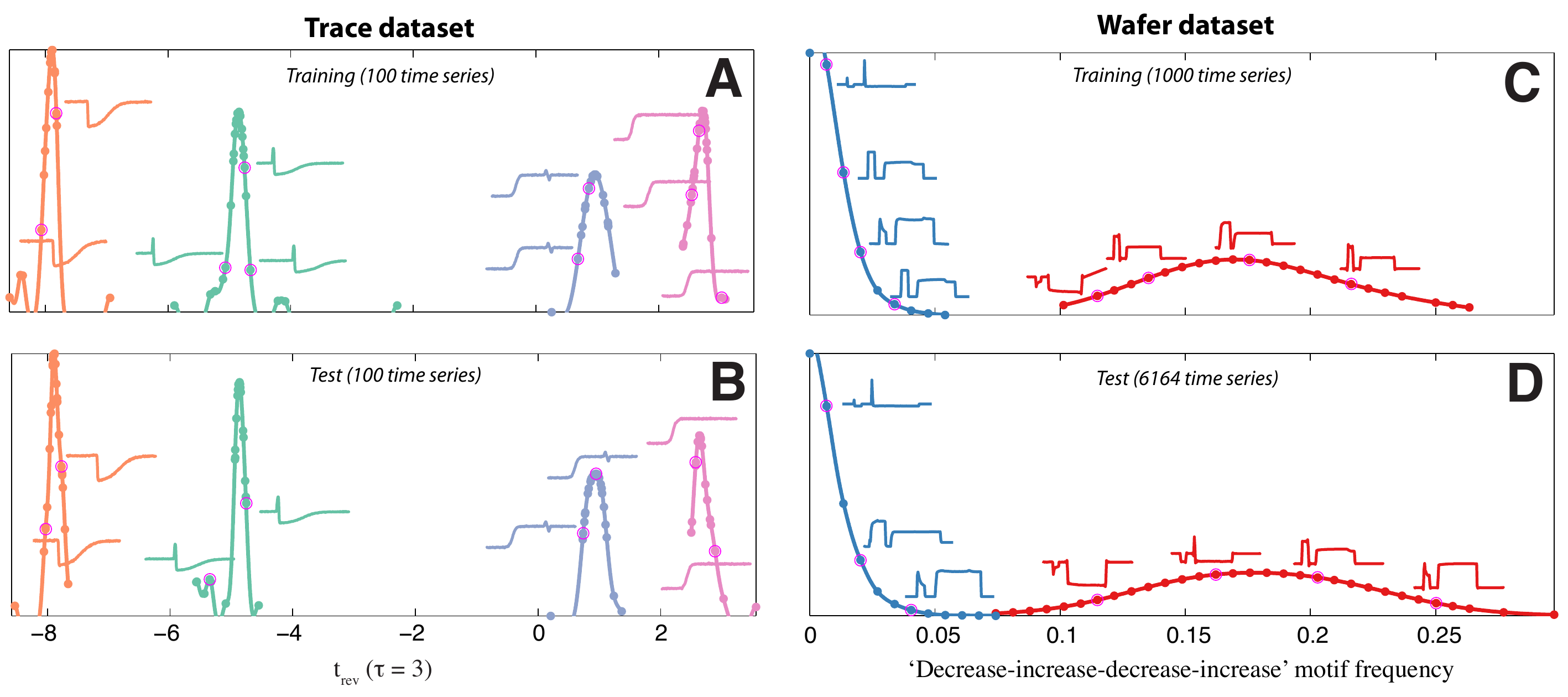}
	\caption{
	\textbf{For some datasets, single extracted features separate the labeled classes accurately.}
	In this way, the dimensionality of the problem is vastly reduced: from the ordered set of $N$ measurements that constitute each time series, to a single feature extracted from it.
	We show two examples: \textbf{A} and \textbf{B} show the feature $t_\mathrm{rev}$ ($\tau = 3$) for the Trace dataset, and \textbf{C} and \textbf{D} plot the proportion of local maxima in time series from the Wafer dataset.
	The distribution of each feature across its range is shown for each of the labeled classes in the training (upper panels) and test (lower panels) sets separately.
	The classes are plotted with different colors, and selected time series (indicated by purple circles) are annotated to each distribution.
	}
	\label{fig:1feat}
\end{figure*}

\subsection{Selected datasets} \label{sec:particular_datasets}

For some datasets, we found that the first selected feature (i.e., $\tilde{f}_1$, the feature with the lowest linear misclassification rate on the training data) distinguished the labeled classes with high accuracy, corresponding to vast dimensionality reduction: from representing time series using all $N$ measured points, to just a single extracted feature.
Examples are shown in Fig.~\ref{fig:1feat} for the Trace and Wafer datasets.
The Trace dataset contains four classes of transients relevant to the monitoring and control of industrial processes \cite{Roverso00}.
There are 25 features in our database that can classify the training set without error, one of which is a time-reversal asymmetry statistic, $t_\mathrm{rev}(\tau = 3)$, where $t_\mathrm{rev}(\tau)$ is defined as
\begin{equation}\label{eqn:DM_trev}
	t_\mathrm{rev}(\tau) = \frac{\langle (x_{t+\tau} - x_{t})^3 \rangle}{\langle (x_{t+\tau} - x_{t})^2 \rangle^{3/2}},
\end{equation}
where $x_i$ are the values of the time series, $\tau$ is the time lag ($\tau = 3$ for this feature), and averages, $\langle\cdot\rangle$, are performed across the time series \cite{Schreiber00}. 
This operation with $\tau = 3$ produces distributions for the four classes of the Trace dataset as shown in Figs.~\ref{fig:1feat}A and B for the training and test sets, respectively.
Simple thresholds on this feature, learned using a linear classifier, allow new time series to be classified by evaluating Eq.~\eqref{eqn:DM_trev}.
In this way, the test set of Trace is classified with 99\% accuracy, producing similar performance as DTW (which classifies the test set without error) but using just a single feature, and circumventing the need to compute distances between pairs of time series.

A second example is shown in Figs.~\ref{fig:1feat}C and \ref{fig:1feat}D for the Wafer dataset, which contains measurements of various sensors during the processing of silicon wafers for semiconductor fabrication that are either `normal' or `abnormal' \cite{Olszweski01}.
As can be seen from the annotations in Figs.~\ref{fig:1feat}C and \ref{fig:1feat}D, each class of time series in this dataset is quite heterogenous.
However, the single feature selected for this dataset simply counts the frequency of the pattern `decrease-increase-decrease-increase' in successive pairs of samples of a time series, expressed as a proportion of the time-series length.
A simple threshold learned on this feature classifies the test set with an accuracy of 99.98\%, slightly higher than the best instance-based result of 99.5\% for Euclidean 1-NN, but much more efficiently: using a single extracted feature rather than comparing all 152 samples of each time series to find matches in the training set.

\begin{figure*}[h]
	\centering
		\includegraphics[width = 16.5cm]{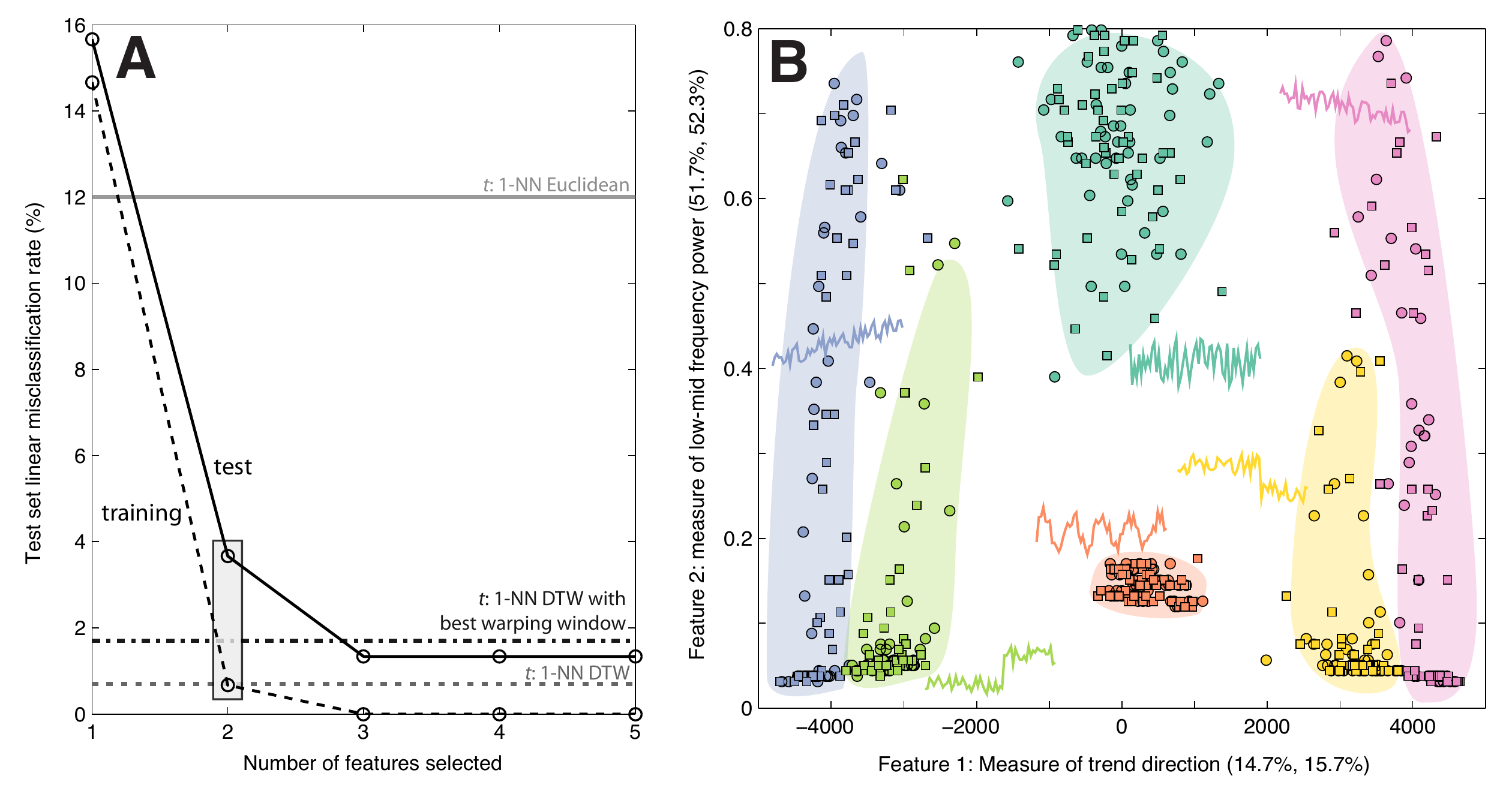}
	\caption{
				\label{fig:SyntheticControlSupervised}
	\textbf{Feature selection for the Synthetic Control dataset.}
	\textbf{A} The training and test set misclassification rates as a function of the number of selected features.
	Our feature selection method terminates after two features (shown boxed) because the subsequent improvement in training set classification rate from adding another feature is less than 3\%.
	Misclassification rates for instance-based classifiers are shown as horizontal lines for comparison (cf. Sec.~\ref{sec:DM_feature_selection}).
	\textbf{B} The training (circles) and test (squares) data are plotted in the space of the two features selected in \textbf{A} (shown boxed), which are described in the main text.
	The test set misclassification rate of each individual feature is indicated in parentheses in the form (training, test).
	The labeled classes: `normal' (dark green), `cyclic' (orange), `increasing trend' (blue), `decreasing trend' (pink), `upward shift' (light green), and `downward shift' (yellow), are well separated in this space.
	A selected time series from each class has been annotated to the plot and background shading has been added manually to guide the eye.
	}
\end{figure*}

Feature-based classifiers constructed for most time-series datasets studied here combine multiple features.
An example is shown in Fig.~\ref{fig:SyntheticControlSupervised} for the Synthetic Control dataset, which contains six classes of noisy control chart patterns, each with distinctive dynamical properties: (i) `normal' (dark green), (ii) `cyclic' (orange), (iii) `increasing trend' (blue), (iv) `decreasing trend' (pink), (v) `upward shift' (light green), (vi) `downward shift' (yellow) \cite{Pham98}.
In statistical process control, it is important to distinguish these patterns to detect potential problems with an observed process.
As shown in Fig.~\ref{fig:SyntheticControlSupervised}A for greedy forward feature selection, the misclassification rate in both the training and test sets drops sharply when a second feature is added to the classifier, but plateaus as subsequent features are added.
The dataset is plotted in the space of these first two selected features, $(\tilde{f}_1,\tilde{f}_2)$, in Fig.~\ref{fig:SyntheticControlSupervised}B.
The first feature, $\tilde{f}_1$, is named \textbf{PH\_ForcePotential\_sine\_10\_004\_10\_median}, and is plotted on the horizontal axis of Fig.~\ref{fig:SyntheticControlSupervised}B.
This feature behaves in a way that is analogous to performing a cumulative sum through time of the $z$-scored time series (the cumulative sum, $S_t$, is defined as $S_t = \sum_{i=1}^t x_i$), and then returning its median (i.e., the median of $S_t$ for $t = 1, 2, ..., N$)\footnote{In fact, this operation treats the time series as a drive to a particle in a sinusoidal potential and outputs the median values of the particle across its trajectory.
However, the parameter values for this operation are such that the force from the potential is so much lower than that of the input drive from the time series that the effect of the sinusoidal potential can be neglected and the result is, to a very good approximation, identical to taking a cumulative sum.}.
This feature takes high values for time series that have a decreasing trend (the cumulative sum of the $z$-scored time series initially increases and then decreases back to zero), moderate values for time series that are approximately mean-stationary (the cumulative sum of the $z$-scored time series oscillates about zero), and low values for time series that have an increasing trend (the cumulative sum of the $z$-scored time series initially decreases and then increases back to zero).
As shown in Fig.~\ref{fig:SyntheticControlSupervised}B, this feature on its own distinguishes most of the classes well, but confuses the two classes without an underlying trend: the uncorrelated random number series, `normal' (green), and the noisy oscillatory time series, `cyclic' (orange).
The second selected feature, $\tilde{f}_2$, named \textbf{SP\_basic\_pgram\_hamm\_power\_q90mel}, is on the vertical axis of Fig.~\ref{fig:SyntheticControlSupervised}B and measures the mel-frequency at which the cumulative power spectrum (obtained as a periodogram using a Hamming window) reaches 90\% of its maximum value\footnote{The `mel scale' is a monotonic transformation of frequency, $\omega$, as $1127\log(\omega/(1400\pi)+1)$.}.
This feature gives low values to the cyclic time series (orange) that have more low-frequency power, and high values to the uncorrelated time series (dark green).
Even though this feature alone exhibits poor classification performance (a misclassification rate of 52.3\% on the test data), it compensates for the weakness of the first feature, which confuses these two classes.
These two features are selected automatically and thus complement one another in a way that facilitates accurate classification of this dataset.
Although DTW is more accurate at classifying this dataset (cf. Fig.~\ref{fig:SyntheticControlSupervised}A), this example demonstrates how selected features can provide an understanding of how the classifier uses interpretable time-series properties to distinguish the classes of a dataset (see Supplementary Fig.~2 for an additional example using the Two Patterns dataset). 
Furthermore, our results follow dimensionality reduction from 60-sample time series down to two simple extracted features, allowing the classifier to be applied efficiently to massive databases and to very long time series (cf. Sec.~\ref{sec:computational_issues}).

\begin{figure}[t]
	\centering
		\includegraphics[width = 7.6cm]{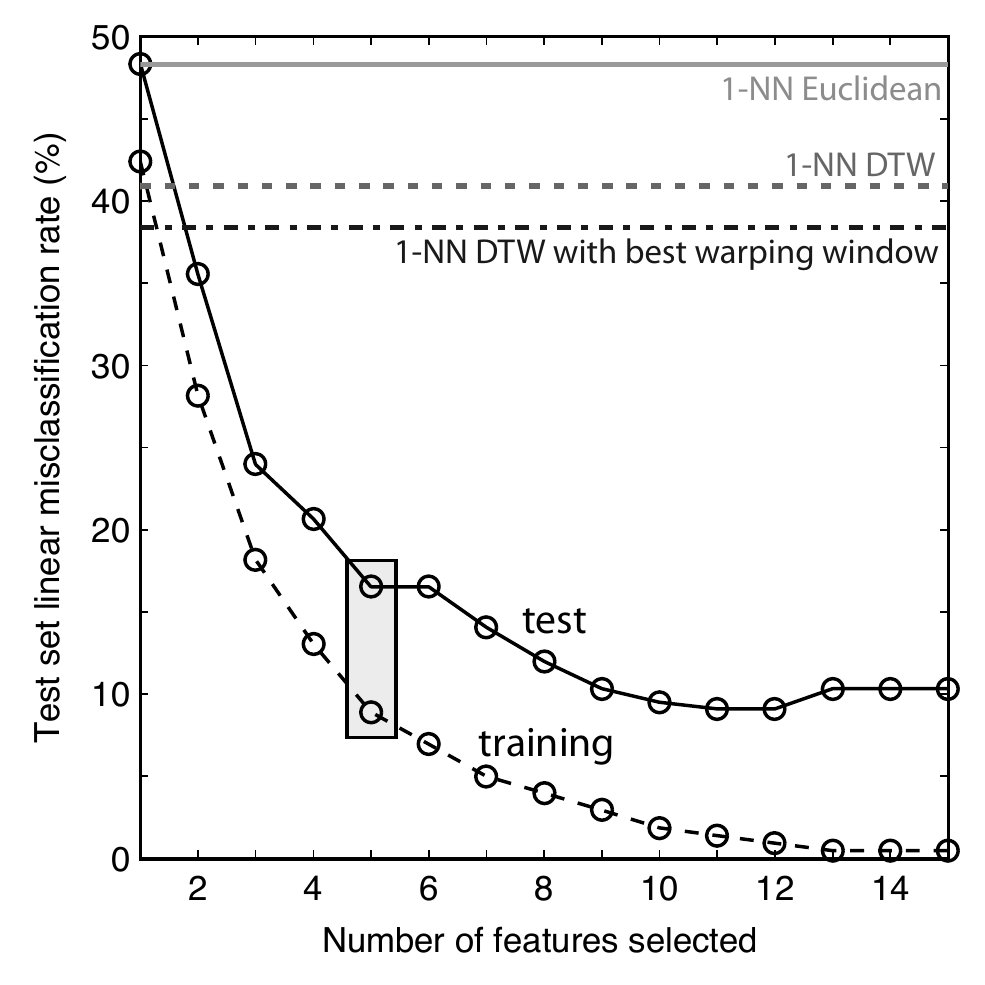}
	\caption{
	\textbf{Feature selection for the OSU Leaf dataset.}
	Training and test set misclassification rates are plotted as a function of the number of features learned using greedy forward feature selection.
	Misclassification rates for three instance-based 1-NN classifiers are shown using horizontal lines for comparison.
	The number of features in the classifier is chosen at the point the improvement in classification rate on adding another feature drops below 3\%, which is five features for this dataset (shown with a gray rectangle).
	}
	\label{fig:OSULeaf}
\end{figure}

For many datasets, such as the six-class OSU Leaf dataset \cite{Gandhi02}, the classification accuracy is improved by including more than two features, as shown in Fig.~\ref{fig:OSULeaf}.
The classification rates of all three 1-NN instance-based classifiers (horizontal lines labeled in Fig.~\ref{fig:OSULeaf}) are exceeded by the linear feature-based classifier with just two features.
The classification performance improves further when more features are added, down to a test set misclassification rate of just 9\% with eleven features (the test set classification rate plateaus as more features are added while the training set classification rate slowly improves, indicating a modest level of over-fitting beyond this point).
The improvement in training-set misclassification rate from adding an additional feature drops below 3\% after selecting five features, yielding a test set misclassification rate of 16.5\% (shown boxed in Fig.~\ref{fig:OSULeaf}), outperforming all instance-based classifiers by a large margin despite dimensionality reduction from 427-sample time series to five extracted features.

\subsection{All results} \label{sec:all_results}
\begin{figure}[t]
	\centering
		\includegraphics[width = 8cm]{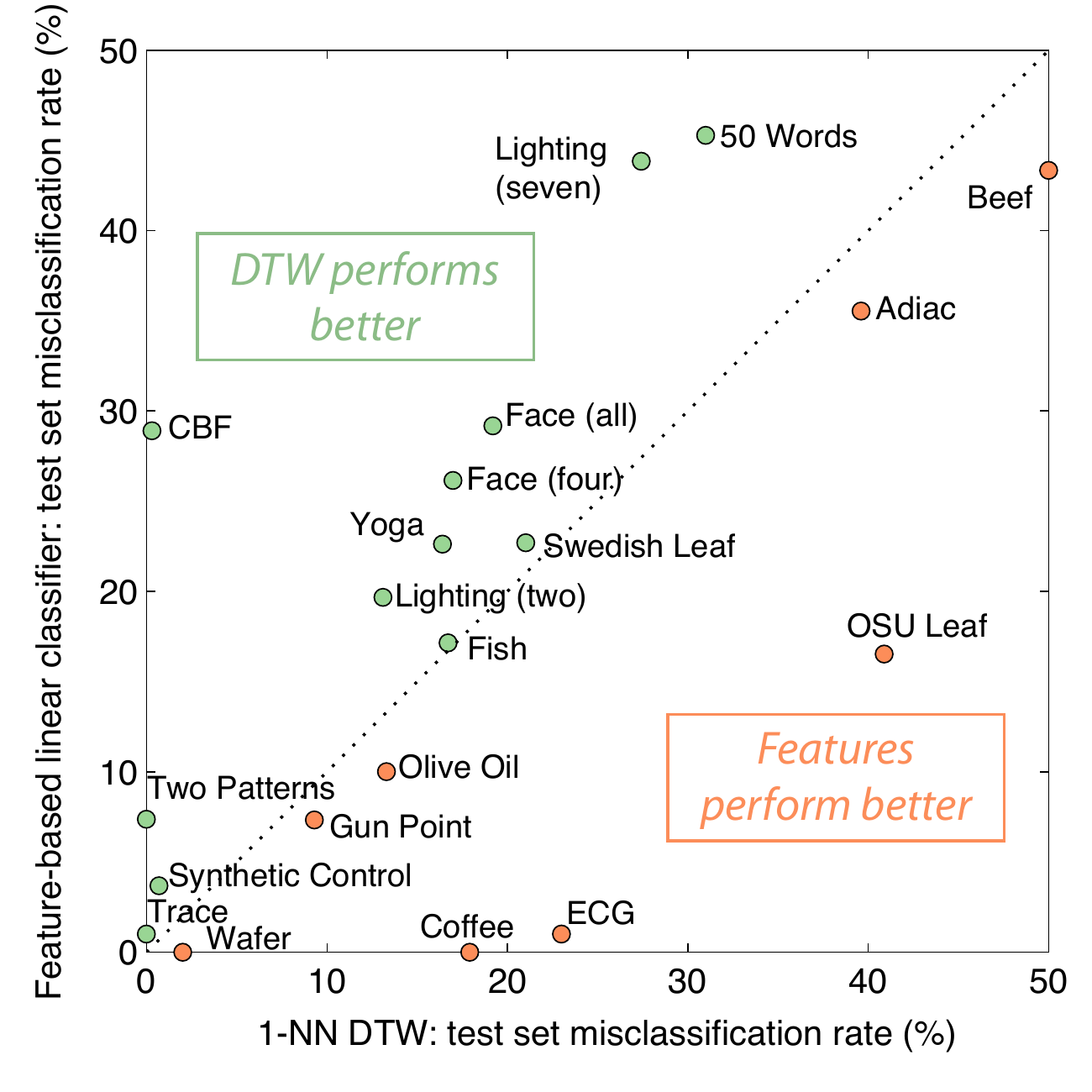}
	\caption{
	\textbf{Test-set misclassification rates of a 1-NN DTW instance-based classifier compared to a feature-based linear classifier.}
	The dotted line indicates the threshold between better performance using features (datasets labeled orange) and using DTW (datasets labeled green).
	Feature-based classification uses an average of $n_\mathrm{feat}$ = 3.2 features compared to an average of $N$ = 282.1 samples that make up each time series (see Table~\ref{tab:cfnresults}).
	}
	\label{fig:compare_performance}
\end{figure}

Having provided some intuition for our method using specific datasets as examples, we now present results for all twenty time-series datasets from {\it The UCR Time Series Classification/Clustering Homepage} (as of mid-2011) \cite{Keogh06a}.
For these datasets of short patterns whose values through time can be used as the basis of computing a meaningful measure of distance between them, DTW has been shown to set a high benchmark for classification performance \cite{Xi06}.
However, as shown above, it is possible for feature-based classifiers to outperform instance-based classifiers despite orders of magnitude of dimensionality reduction.
Results for all datasets are shown in Table~\ref{tab:cfnresults}, including test set misclassification rates for three instance-based classifiers and for our linear feature-based classifier.
The final two columns of Table~\ref{tab:cfnresults} demonstrate extensive dimensionality reduction using features for all datasets, using an average of $n_\mathrm{feat} = $ 3.2 features to represent time series containing an average of $N = $ 282.1 samples.
A direct comparison of 1-NN DTW with our linear feature-based classifier is shown in Fig.~\ref{fig:compare_performance} for all datasets.
Both methods yield broadly similar classification results for most datasets, but some datasets exhibit large improvements in classification rate using one method over the other.
Note that results showing the variation across different training/test partitions (retaining training/test proportions) are shown in Supplementary Table~II for Euclidean 1-NN and our linear feature-based classifier; results are mostly similar as shown here for fixed partitions. 
Across all datasets, a wide range of time-series features were selected for classification, including measures of autocorrelation and automutual information, motifs in symbolized versions of time series, spectral properties, entropies, measures of stationarity, outlier properties, scaling behavior, and others.
Names of all features selected for each dataset, along with their Matlab code names, are provided in Supplementary Table~III. 

\begin{table*}[h]
\caption{
\textbf{Misclassification rates for classifiers applied to all twenty time-series datasets analyzed in this work.}
Results are shown for three instance-based 1-NN classifiers, and our linear feature-based classifier, labeled `Feature-based linear', in which time series are represented using a set of extracted features, trained by greedy forward feature selection using a linear classifier from a database containing thousands of features.
All percentages in the table are test set misclassification rates.
Results using dynamic time warping (DTW) were obtained from {\it The UCR Time Series Classification/Clustering Homepage} \cite{Keogh06a}.
`Warping window' has been abbreviated as `WW' in the third column, and the parameter $r$ is expressed as a percentage of the time-series length.
The classifier with the lowest misclassification rate for each dataset is printed in boldface.
The final two columns list the number of samples, $N$, in each time series in the dataset, that are used in instance-based classification, compared to the number of features, $n_\mathrm{feat}$, chosen from feature selection.
The dimensionality reduction from using feature selection is large, while classification performance is often comparable or superior (cf. Fig.~\ref{fig:compare_performance}).
Note that results for different training/test partitions of the datasets are mostly similar to those shown here, and are in Supplementary Table~II. 
}
\small
\centering
\begin{tabular}{|c||c|c|c||c||c|c|}
\hline
Dataset & Euclidean & DTW & DTW 1-NN & Feature-based, & $N$ (samples) & $n_\mathrm{feat}$ \\
& 1-NN (\%) & 1-NN (\%) & best WW [$r$] (\%) & linear (\%) & & \\
\hline
{\it Synthetic Control}	 & 12.0 & \textbf{0.7} & 1.7 [6] & 3.7 & 60 & 2 \\ 
\hline                                              
{\it Gun point} 		 & 8.7  & 9.3  & 8.7 [0] & \textbf{7.3} & 150 & 2 \\ 
\hline                                              
{\it CBF} 				 & 14.8 & \textbf{0.3} & 0.4 [11] & 28.9 & 128 & 2 \\ 
\hline                                              
{\it Face (all)} 		 & 28.6 & \textbf{19.2} & \textbf{19.2} [3] & 29.2 & 131 & 5 \\ 
\hline                                              
{\it OSU Leaf} 			 & 48.3 & 40.9 & 38.4 [7] & \textbf{16.5} & 427 & 5 \\ 
\hline                                              
{\it Swedish Leaf} 		 & 21.1 & 21.0 & \textbf{15.7} [2] & 22.7 & 128 & 5 \\ 
\hline                                              
{\it 50 Words} 			 & 36.9 & 31.0 & \textbf{24.2} [6] & 45.3 & 270 & 7 \\ 
\hline                                              
{\it Trace} 			 & 24.0 & \textbf{0.0}  & 1.0 [3] & 1.0 & 275 & 1 \\ 
\hline                                              
{\it Two Patterns} 		 & 9.3  & \textbf{0.0} & 0.2 [4] & 7.4 & 128 & 2 \\ 
\hline                                              
{\it Wafer} 			 & 0.5  & 2.0  & 0.5 [1] & \textbf{0.0} & 152 & 1 \\ 
\hline                                              
{\it Face (four)}		 & 21.6 & 17.0 & \textbf{11.4} [2] & 26.1 & 350 & 3 \\ 
\hline                                              
{\it Lightning (two)} 	 & 24.6 & \textbf{13.1} & \textbf{13.1} [6] & 19.7 & 637 & 2 \\ 
\hline                                              
{\it Lightning (seven)}  & 42.5 & \textbf{27.4} & 28.8 [5] & 43.8 & 319 & 4 \\ 
\hline                                              
{\it ECG}				 & 12.0 & 23.0 & 12.0 [0] & \textbf{1.0} & 96 & 1 \\ 
\hline                                              
{\it Adiac}				 & 38.9 & 39.6 & 39.1 [3] & \textbf{35.5} & 176 & 5 \\ 
\hline                                              
{\it Yoga}				 & 17.0 & 16.4 & \textbf{15.5} [2] & 22.6 & 426 & 3 \\ 
\hline                                              
{\it Fish}				 & 21.7 & 16.7 & \textbf{16.0} [4] & 17.1 & 463 & 6 \\ 
\hline                                              
{\it Beef}				 & 46.7 & 50.0 & 46.7 [0] & \textbf{43.3} & 470 & 5 \\ 
\hline                                              
{\it Coffee}			 & 25.0 & 17.9 & 17.9 [3] & \textbf{0.0} & 286 & 1 \\ 
\hline                                              
{\it Olive Oil}			 & 13.3 & 13.3 & 16.7 [1] & \textbf{10.0} & 570 & 2 \\ 
\hline
\end{tabular}
\label{tab:cfnresults}
\end{table*}

As with all approaches to classification, a feature-based approach is better suited to some datasets than others \cite{Wolpert97}.
Indeed, we found that feature-based classifiers outperform instance-based alternatives on a number of datasets, and sometimes by a large margin.
For example, in the ECG dataset, the feature-based classifier yields a test set misclassification rate of 1.0\% using just a single extracted feature, whereas the best instance-based classifiers (Euclidean 1-NN and DTW 1-NN using the best warping window) have a misclassification rate of 12.0\%.
In the Coffee dataset, the test set is classified without error using a single extracted feature, whereas the best instance-based classifiers (both using DTW) have a misclassification rate of 17.9\%.
In other cases, instance-based approaches (including even the straightforward Euclidean 1-NN classifier) performed better.
For example, the 50 Words dataset has a large number of classes (fifty) and a large heterogeneity in training set size (from as low as 1 to 52 training examples in a given class), for which matching to a nearest neighbor using instance-based methods outperforms the linear feature-based classifier.
The Face (four) dataset also has relatively few, quite heterogenous, and class-unbalanced training examples, making it difficult to select features that best capture the class differences; instance-based methods also outperform our feature-based approach on this dataset.
The ability of DTW to adapt on a pairwise basis to match each test time series to a similar training time series can be particularly powerful mechanism for some datasets, and is unavailable to a static, feature-based classifier, which does not have access to the training data once the classifier has been trained.
This mechanism is seen to be particularly important for the Lightning (seven) dataset, which contains heterogenous classes with unaligned patterns---DTW performs well here (misclassification rate of 27.2\%), while 1-NN Euclidean distance and feature-based classifiers perform worse, with misclassification rates exceeding 40\%.
Our feature-based classifiers are trained to optimize the classification rate in the training set, and thus assume similar class proportions in the test set, which is often not the case; by simply matching instances of time series to the training set, discrepancies between class ratios in training and test sets are less problematic for instance-based classification.
This may be a contributing factor to the poor performance of feature-based classification for the 50 Words, Lightning (seven), Face (four) and Face (all) datasets.
A feature-based representation also struggles when only a small number of heterogenous training examples are available, as with the 50 Words, Lightning (seven), and CBF datasets.
In this case it can be difficult to select features that represent differences within a class as `the same', and simultaneously capture differences between classes as `different'.
Although we demonstrate improved performance on the Adiac dataset, with a misclassification rate of 35.5\%, this remains high.
We note that the properties of this dataset provide multiple challenges for our method, that may also contribute to its difficulty with instance-based approaches, including a small number and large variation in the number of examples in the training set (between 5 and 15 examples per class), a negative correlation between training set size and test set size (where our method assumes the same class proportions in the test set), and a large number of classes (37), which are relatively heterogenous within a given class, and visually quite similar between classes.

Despite some of the challenges of feature-based classification, representing time series using extracted features brings additional benefits, including vast dimensionality reduction and, perhaps most importantly, interpretable insights into the differences between the labeled classes (as demonstrated in Sec.~\ref{sec:particular_datasets}).
This ability to learn about the properties and mechanisms underlying class differences in the time series in some sense corresponds to the `ultimate goal of knowledge discovery' \cite{Morchen03}, and provides a strong motivation for pursuing a feature-based representation of time-series datasets where appropriate.

\subsection{Computational complexity} \label{sec:computational_issues}
In this section, the computational effort required to classify time series using extracted features is compared to that of instance-based approaches.
Calculating the Euclidean distance between two time series has a time complexity of $\mathcal{O}(N)$, where $N$ is the length of each time series (which must be constant).
The distance calculation for dynamic time warping (DTW) has a time complexity of $\mathcal{O}(N^2)$ in general, or $\mathcal{O}(Nw)$ using a warping window, where $w$ is the warping window size \cite{Wang12}.
Classifying a new time series using a 1-NN classifier and sequential search (i.e., sequentially calculating distances between a new time series and all time series in the training set) therefore has a time complexity of $\mathcal{O}(N n_\mathrm{train})$ for Euclidean distances and either $\mathcal{O}(N^2 n_\mathrm{train})$ or $\mathcal{O}(N w n_\mathrm{train})$ for DTW, where $n_\mathrm{train}$ is the number of time series in the training set \cite{Wang12}.
Although the amortized time complexity of the distance calculation can be improved using lower bounds \cite{Rakthanmanon12, Wang12, Keogh03}, and speedups can be obtained using indexing \cite{Ding08, Chakrabarti02, Shieh08} or time-domain dimensionality reduction \cite{Lin07}, the need to calculate many distances between pairs of time series is fundamental to instance-based classification, such that scaling with the time-series length, $N$, and the size of the training set, $n_\mathrm{train}$, is inevitable.
While the use of shapelets \cite{Ye09, Rakthanmanon13} addresses some of these issues, here we avoid comparisons in the time domain completely and instead classify time series using a static representation in terms of extracted features.
Time-domain classification can therefore become computationally prohibitive for long time series and/or very large datasets.

In contrast to instance-based classifiers, the bulk of the computational burden for our feature-based method is associated with learning a classification rule on the training data, which involves the computation of thousands of features for each training time series, which can be lengthy.
However, this is a one-off computational cost: once the classifier has been trained, new time series are classified quickly and independent of the training data.
For most cases in this work, selected features correspond to simple algorithms with a time complexity that scales linearly with the time series length, as $\mathcal{O}(N)$.
The classification of a new time series then involves simply computing $n_\mathrm{feat}$ features, and then evaluating the linear classification rule.
Hence, if all features have time complexities that scale as $\mathcal{O}(N)$, the total time complexity of classifying a new time series scales as $\mathcal{O}(N n_\mathrm{feat})$ if the features are calculated serially (we note, of course, that calculating each of the $n_\mathrm{feat}$ features can be trivially distributed).
This result is independent of the size of the training dataset and, importantly, the classification process does not require any training data to be loaded into memory, which can be a major limitation for instance-based classification of large datasets.

\begin{figure}[t]
	\centering
		\includegraphics[width = 9cm]{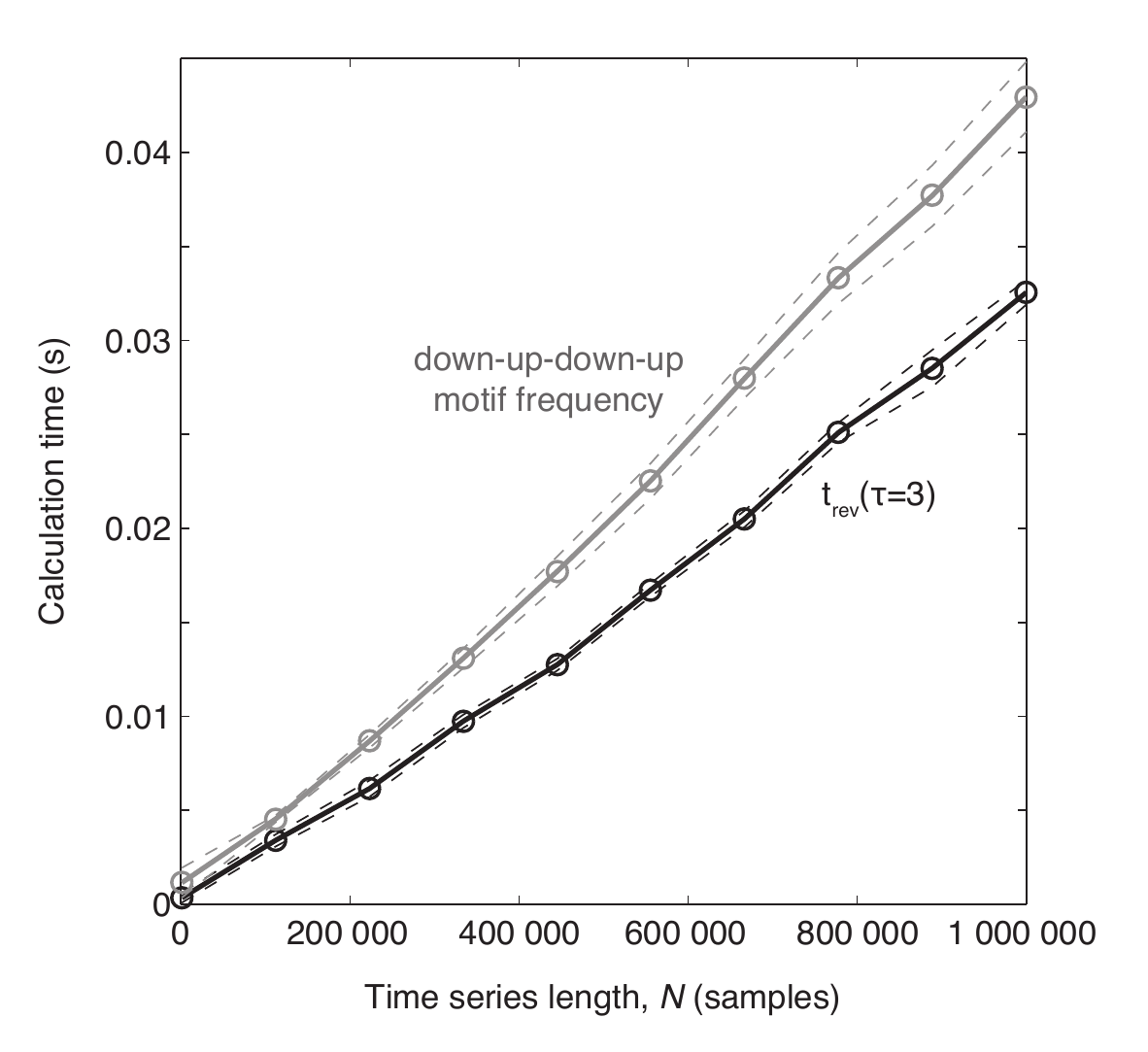}
	\caption{\textbf{Scaling of two features with time-series length, $N$}.
	The features are those selected for (i) the Trace dataset: $t_\mathrm{rev}(\tau=3)$, and (ii) the Wafer dataset: the normalized frequency of the decrease-increase-decrease-increase motif (labeled `down-up-down-up motif frequency' in the plot), cf. Fig.~\ref{fig:1feat}.
	Both operations have a time complexity that scales approximately linearly with the time-series length, $N$.
	Calculation times were evaluated on Gaussian-distributed white noise time series and each point represents the average of 100 repeats of the calculation on a basic desktop PC, along with a standard deviation either side of the mean.
	Unlike instance-based classification that requires the calculation of distances between many pairs of time series, this feature-based approach can be applied to much longer time series, where the time taken to reduce a time series of 100\,000 samples, for example, to one of these relevant features, is of the order of milliseconds.
	}
	\label{fig:timescaling}
\end{figure}

Having outlined the computational steps involved in feature-based classification, we now describe the actual time taken to perform classification using specific examples.
First we show that even though the methods used in this work were applied to relatively short time series (of lengths between 60 and 637 samples), they are also applicable to time series that are orders of magnitude longer (indeed many operations are tailored to capturing complex dynamics in long time-series recordings).
For example, the features selected for the Trace and Wafer datasets shown in Fig.~\ref{fig:1feat} were applied to time series of different lengths, as plotted in Fig.~\ref{fig:timescaling}.
Note that the following is for demonstration purposes only: these algorithms were implemented directly in Matlab and run on a normal desktop PC with no attempt to optimize performance.
The figure shows that both of these operations have a time complexity that scales approximately linearly with the time-series length, as $\mathcal{O}(N)$.
Feature-based classification is evidently applicable to time series that are many orders of magnitude longer than short time-series patterns (as demonstrated in previous work \cite{Fulcher13})---in this case a 100\,000-sample time series is converted to a single feature: either $t_\mathrm{rev}(\tau = 3)$, or the decrease-increase-decrease-increase motif frequency, in under 5\,ms.
Note that although simple $\mathcal{O}(N)$ operations tended to be selected for many of the datasets studied in this work, other more sophisticated operations (those based on nonlinear model fits, for example) have computational time complexities that scale nonlinearly with the time-series length, $N$.
The time complexity of any particular classifier thus depends on the features selected in general (however, in future computational constraints could be placed on the set of features searched across, e.g., restricting the search to features that scaled linearly as $\mathcal{O}(N)$, as discussed in Sec~\ref{sec:discussion}).

Next we outline the sequence of calculations involved in classifying the Wafer dataset as a case study.
We emphasize that in this paper, we are not concerned with optimizing the one-off cost of training a classifier and simply calculated the full set of 9\,288 features on each training dataset, despite high levels of redundancy in this set of features \cite{Fulcher13} and the inclusion of thousands of nonlinear methods designed for long streams of time-series data.
In future, calculating a reduced set (of say 50 features) could reduce the training computation times reported here by orders of magnitude.
The calculation of this full set of 9\,288 features on a (152-sample) time series from the Wafer dataset took an average of approximately 31\,s.
Performing these calculations serially for this very large training set with $n_\mathrm{train} =$ 1\,000, this amounts to a total calculation time of 8.6 hours.
This is the longest training time of any dataset studied here due to a large number of training examples; other datasets had as few as 24 training examples, with a total training time under 15\,min.
Furthermore, all calculations are independent of one another and can be trivially distributed; for example, with as many nodes as training time series, the total computation is the same as for a single time series, $\sim$30\,s in this case (or, furthermore, with as many nodes as time series/operation pairs, the total computation time is equal to that of the slowest single operation operating on any single time series, reducing the computation time further).
For the Wafer dataset, feature selection took 6\,s, which produced a (training set) misclassification rate of 0\% and terminated the feature selection process.
Although just a single feature was selected here, more features are selected in general, which take $\sim$ 6--10\,s per feature to select.
It then took a total of 32.5\,s to load all 6\,164 test time series into memory, a total of 0.1\,s to calculate the selected feature and evaluate the linear classification rule on all time series on a basic desktop PC.
The result classified 6\,163 of the 6\,164, or 99.98\%, of the test time series correctly.

In summary, the bulk of the computational burden of our highly comparative feature-based classification involves the calculation of thousands of features on the training data (which could be heavily optimized in future work).
Although instance-based methods match new time series to training instances and do not require such computational effort to train, the investment involved in training a feature-based classification rule allows new time series to be classified rapidly and independent of the training data.
The classification of a new time series simply involves extracting feature(s) and evaluating a linear classification rule, which is very fast ($\approx$ 3$\times$10$^{-8}$\,s per time series for the Wafer example above), and limited by the loading of data into memory ($\approx$ 5$\times$10$^{-3}$\,s per time series for the Wafer dataset).
In general, calculation times will depend on the time-series length, $N$, the number of features selected during the feature selection process, $n_\mathrm{feat}$, and the computational time complexity of those selected features.
Performing feature-based classification in this way is thus suited to applications that value fast, real-time classification of new time series and can accommodate the relatively lengthy training process (or where sufficient distributed computational power is available to speed up the training process).
There are clear applications to industry, where measured time series need to be checked in real time on a production line, for quality control, or the rapid classification of large quantities of medical samples, for example.

%

\section{Discussion} \label{sec:discussion}
In summary, we have introduced a highly comparative method for learning feature-based classifiers for time series.
Our main contributions are as follows:\\
(i) Previous attempts at feature-based time-series classification in the data mining literature have reported small sets of ($\sim$10 or fewer) manually selected or generic time-series features.
Here, for the first time, we apply a diverse set of thousands of time-series features and introduce a method that compares across these features to construct feature-based classifiers automatically.\\
(ii) Features selected for the datasets studied here included measures of outlier properties, entropy measures, local motif frequencies, and autocorrelation-based statistics.
These features provide interpretable insights into the properties of time series that differ between labeled classes of a dataset.\\
(iii) Of the twenty UCR time-series datasets studied here, feature-based classifiers used an average of 3.2 features compared to an average time-series length of 282.1 samples, representing two orders of magnitude of dimensionality reduction.\\
(iv) Despite dramatic dimensionality reduction and an inability to compare and match similar patterns through time, our feature-based representations of time series produced good classification performance that was in many cases superior to DTW, and in some cases by a large margin.\\
(v) Unlike instance-based classification, the training of a feature-based classifier incurs a significant computational expense.
However, this one-off cost allows new time series to be classified extremely rapidly and independent of the training set. Furthermore, there is much scope for optimizing this training cost in future by exploiting redundancy in our massive feature set.


To introduce the highly comparative approach, we have favored the interpretability of feature selection and classification methods over their sophistication.
Feature selection was achieved using greedy forward selection, and classification was done using linear discriminant classifiers.
Many more sophisticated feature selection \cite{Jain00, Guyon03, Guyon07} and classification \cite{Hastie09} methods exist (e.g., that allow for more robust and/or nonlinear classification boundaries) and should improve the classification results presented here.
This flexibility to incorporate a large and growing literature of sophisticated classifiers operating on feature vectors, including decision trees and support vector machines or even $k$-NN applied in a feature space, is a key benefit of our approach \cite{Hastie09}.
Considering combinations of features that are not necessarily the result of a greedy selection process (e.g., classifiers that combine features with poor individual performance have been shown to be very powerful on some datasets \cite{Guyon03}), should also improve classification performance.
However, we note that complex classifiers may be prone to over-fitting the training data and thus may require cross-validation on the training data to reduce the in-sample bias.
However, cross-validation is problematic for some of the datasets examined here that have small numbers of training examples (as low as just a single training example for a class in the 50 Words dataset).
We used the total classification rate as a cost function for greedy forward feature selection to aid comparison to other studies, even though many datasets have unequal numbers of time series in each class and different class proportions in the training and test sets, thus focusing the performance of classifiers towards those classes containing the greatest number of time series.
In future, more subtle cost functions could be investigated, that optimize the mean classification rate across classes, for example, rather than the total number of correct classifications.
In summary, the simple classification and feature selection methods used here were chosen to demonstrate our approach as clearly as possible and produce easily-interpretable results; more sophisticated methods could be investigated in future to optimize classification accuracies for real applications.

Because we used thousands of features developed across many different scientific disciplines, many sets of features are highly correlated to one another \cite{Fulcher13}.
Greedy forward feature selection chooses features incrementally based on their ability to increase classification performance, so if a feature is selected at the first iteration, a highly correlated feature is unlikely to increase the classification rate further.
Thus, the non-independence of features does not affect our ability to build successful feature-based classifiers in this way.
However, strong dependencies between operations can mean that features selected using different partitions of the data into training and testing portions can be different (or even for the same partition when two or more features yield the same classification rate and are selected at random).
For homogenous datasets, features that differ for different data partitions are typically slight variants of one another; for example, the second feature selected for the Synthetic Control dataset (cf. Sec.~\ref{sec:particular_datasets}) is a summary of the power spectrum for some partitions and an autocorrelation-based measure for others---both features measure aspects of the linear correlation present in the time series and thus contribute a similar understanding of the time series properties that are important for classification.
The selection of either feature yields similar performance on the unseen data partition.
We also note that this redundancy in the feature set could be exploited in future to produce a powerful reduced set of approximately independent, computationally inexpensive, and interpretable features with which to learn feature-based classifiers for time series.
Future work could also focus on adding new types of features found to be useful for time-series classification (or comparing them to our implementation of existing methods, cf. \cite{Fulcher13}), as our ability to construct useful feature-based time-series classifiers is limited by those features contained in our library of features, which is currently comprehensive but far from exhaustive.
Together, these refinements of the feature set could dramatically speed up the computation times reported here, improve the interpretability of selected features, and increase classification performance.

Many features in our database are designed for long, stationary streams of recorded data and yet here we apply them to short and often non-stationary time series.
For example, estimating the correlation dimension of a time-delay embedded time series requires extremely long and precise recordings of a system \cite{Kantz04}.
Although the output of a correlation dimension estimate on a 60-sample time series will not be a robust nor meaningful estimate of the correlation dimension, it is nevertheless the result of an algorithm operating on a time series and may still contain some useful information about its structure.
Regardless of the conventional meaning of a time-series analysis method therefore, our approach judges features according to their demonstrated usefulness in classifying a dataset.
Appropriate care must therefore be taken in the interpretation of features should they prove to be useful for classifying a given dataset.

Although feature-based and instance-based approaches to time-series classification have been presented as opposing methodologies here, future work could link them together.
For example, Batista et al. \cite{Batista11} used a simple new feature claimed to resemble `complexity', to rescale conventional Euclidean distances calculated between time series, demonstrating an improvement in classification accuracy.
Rather than using this specific, manually-selected feature, our highly comparative approach could be used to find informative but computationally inexpensive features to optimally rescale traditional Euclidean distances.

\section{Conclusions}\label{sec:conclusions}
In 1993, Timmer et al. \cite{Timmer93} wrote: ``The crucial problem is not the classificator function (linear or nonlinear), but the selection of well-discriminating features. In addition, the features should contribute to an understanding [...].''
In this work, we applied an unprecedented diversity of scientific time-series analysis methods to a set of classification problems in the temporal data mining literature and showed that successful classifiers can be produced in a way that contributes an understanding of the differences in properties between the labeled classes of time series.
Although the datasets studied here are well suited to instance-based classification, we showed that a highly comparative method for constructing feature-based representations of time series can yield competitive classifiers despite vast dimensionality reduction.
Relevant features and classification rules are learned automatically from the labeled structure in the dataset, without requiring any domain knowledge about how the data were generated or measured, allowing classifiers to adapt to the data, rather than attempting to develop classifiers that work `best' on generic datasets.
Although the computation of thousands of features can be intensive (if not distributed), once the features have been selected and the classification rule has been learned, the classification of new time series is rapid and can outperform instance-based classification.
The approach can be applied straightforwardly to time series of variable length, and to time series that are many orders of magnitude longer than those studied here.
Perhaps most importantly, the results provide an understanding of the key differences in properties between different classes of time series, insights that can guide further scientific investigation.
The code for generating the features used in this work is freely available at \url{http://www.comp-engine.org/timeseries/}.


%

%

\ifCLASSOPTIONcompsoc
  \section*{Acknowledgments}
\else
  \section*{Acknowledgment}
\fi
The authors would like to thank Sumeet Agarwal for helpful feedback on the manuscript.

\ifCLASSOPTIONcaptionsoff
  \newpage
\fi

\end{document}